\documentclass[10pt,twocolumn,letterpaper]{article}

\usepackage{iccv}
\usepackage{times}
\usepackage{epsfig}
\usepackage{graphicx}
\usepackage{amsmath}
\usepackage{amssymb}
\usepackage{xargs}
\usepackage{adjustbox}
\usepackage{xcolor}
\usepackage{amsmath}
\usepackage{algorithm}
\usepackage{authblk}
\usepackage[noend]{algpseudocode}

\usepackage[breaklinks=true,bookmarks=false]{hyperref}

\usepackage{color, colortbl}
\definecolor{HighLight}{rgb}{0.88,1,1}
\definecolor{Gray}{gray}{0.9}
\usepackage{caption} 
\newcommand\Tstrut{\rule{0pt}{2.6ex}}         

\iccvfinalcopy 

\def\iccvPaperID{****} 

\begin{document}

\title{Many Task Learning with Task Routing}
\author{Gjorgji Strezoski, Nanne van Noord and Marcel Worring\\University of Amsterdam \\ \texttt{\{g.strezoski, n.j.e.vannoord, m.worring\}@uva.nl}}

\maketitle

\begin{abstract}
	Typical multi-task learning (MTL) methods rely on architectural adjustments and a large trainable parameter set to jointly optimize over several tasks. However, when the number of tasks increases so do the complexity of the architectural adjustments and resource requirements. In this paper, we introduce a method which applies a conditional feature-wise transformation over the convolutional activations that enables a model to successfully perform a large number of tasks. To distinguish from regular MTL, we introduce Many Task Learning (MaTL) as a special case of MTL where more than 20 tasks are performed by a single model. Our method dubbed Task Routing (TR) is encapsulated in a layer we call the Task Routing Layer (TRL), which applied in an MaTL scenario successfully fits hundreds of classification tasks in one model. We evaluate our method on 5 datasets against strong baselines and state-of-the-art approaches. 
\end{abstract}

\section{Introduction}

Multi-tasking is ubiquitous. In everyday life, as well as in computer science, performing multiple tasks at the same time improves efficiency and resource utilization \cite{behavioral_science, xue2007multi}. By definition, MTL is a learning paradigm that seeks to improve the generalization performance of machine learning models by optimizing for more than one task simultaneously \cite{Caruana1997}. Its counterpart, Single Task Learning (STL) occurs when a model is optimized for performing a single task only. STL models usually have an abundance of parameters that have the capacity to fit to more than one task \cite{li2018learning}. In MTL the aim is to make use of this extra capacity. The simultaneously performed tasks can either help or hurt each others execution by sharing the expertise the model develops for each of them during training. For example, in a dataset of bird images, training a model to recognize white head feathers and white underbelly can help in classifying a Seagull bird. However, we cannot expect a bird size detector to help with Seagull classification as this bird appears in many sizes in nature, so size is not relevant to its species.

\begin{figure}
	\includegraphics[width=\linewidth]{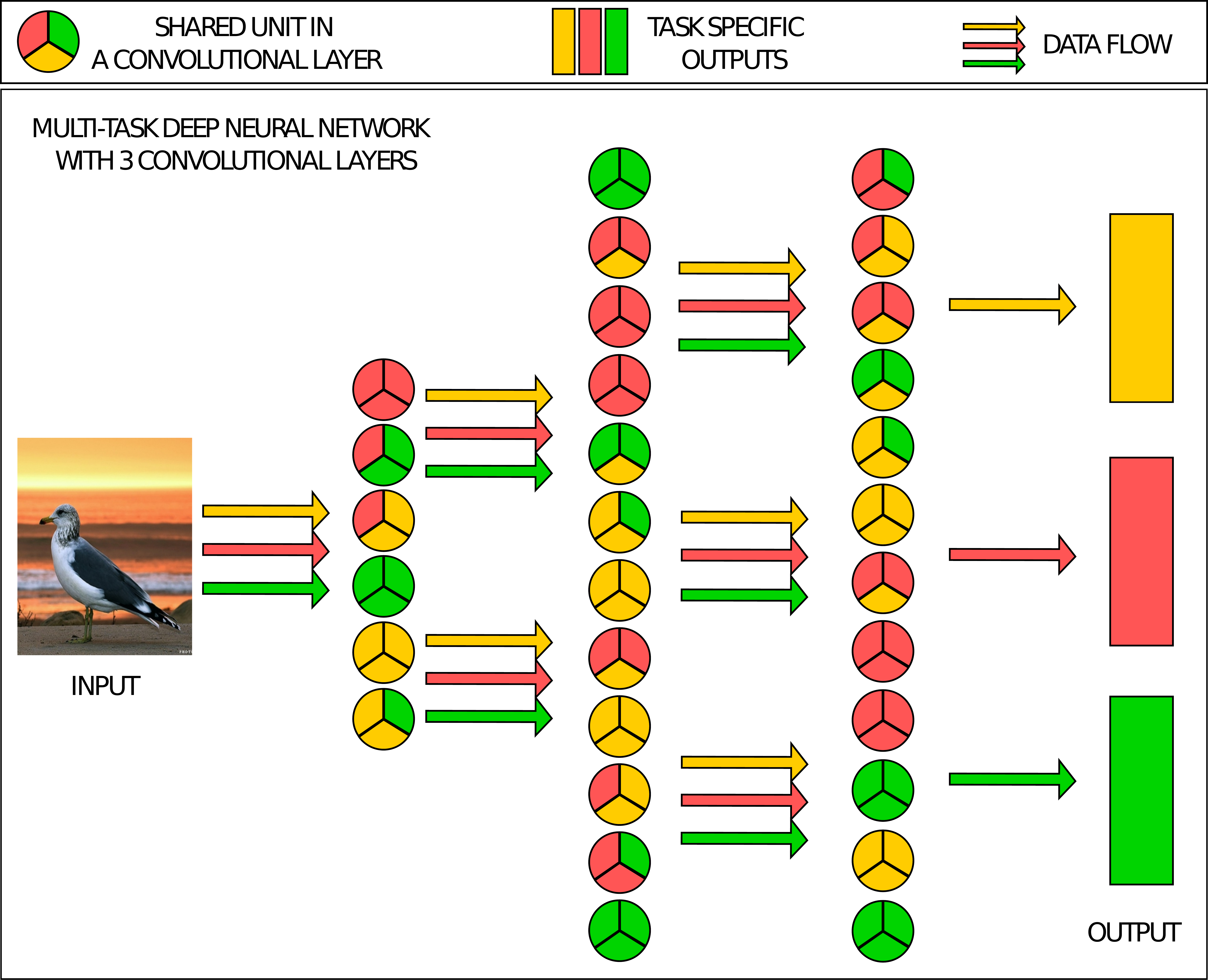}
	\caption{Routing map and specialized subnets through a three layer multi-task deep convolutional neural network with 50\% of the units being shared per task routing layer. \label{fig_one}}
\end{figure}

As with any combinatorial problem, in MTL there exists an optimal combination of tasks and shared resources which is unknown. Searching the space to find this combination is becoming increasingly inefficient, as modern models \cite{he2016deep, huang2017densely, shazeer2017outrageously, real2018regularized} grow in depth, complexity and capacity. This search duration grows proportionally with the number of tasks and parameters present in the model's structure. Previous works in both MTL and STL rely on evolutionary algorithms \cite{architecture_search_mtl} or factorization techniques \cite{yang2017deep} to discover their optimal way of learning, however this takes time and prolongs the training process. In our work, inspired by the efficiency of Random Search \cite{bergstra2012random} we enforce a structured random solution to this problem by regulating the per-task data-flow in our models. As depicted in Figure \ref{fig_one}, by assigning each unit to a subset of tasks that can use it we create specialized sub-networks for each task. In addition, we show that providing tasks with alternate routes through the parameter space, increases feature robustness and improves scalability while boosting or maintaining predictive performance. 

Creating and keeping alternate task routes throughout training, accounts for more than just learning a robust shared and task-specific feature representation. It addresses the problem of \textit{destructive interference} in MTL introduced by X. Zhao et al. \cite{modulation2018}. The authors describe it as the negative influence between tasks that share a representation. In our work, by only sharing partial structures between tasks we distribute the destructive interference effects. This way the negative effects can be diminished in the joint optimization process with tasks that provide a positive influence. In addition, enforcing the defined sharing structure during training, allows for our models to adapt to and overcome destructive interference, if it is present.

\begin{figure*}
	\includegraphics[width=\textwidth]{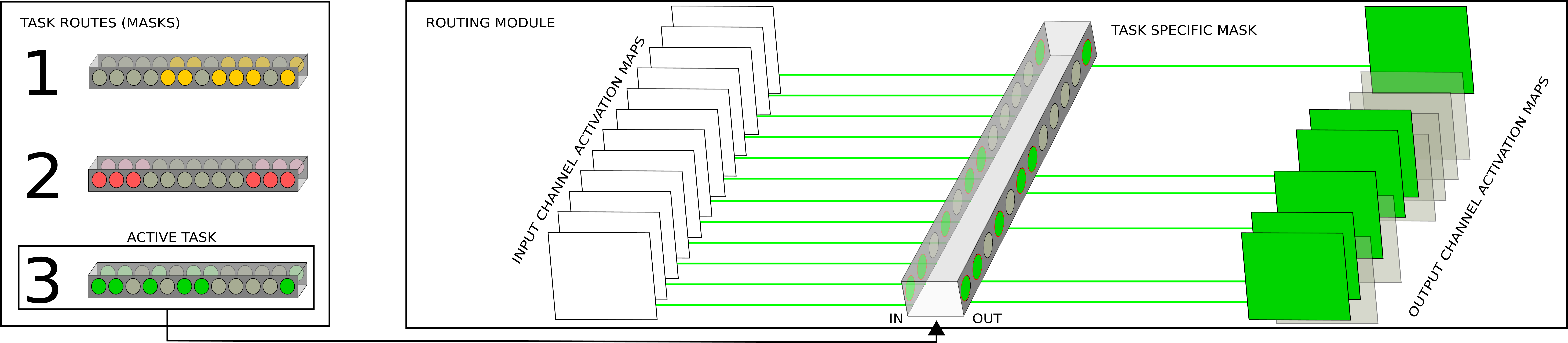}
	\caption{Operation of the TRL over the output from a convolutional layer (white channels). The current active task is used to select the mask (bright green are 1, and dark green are 0). After the element-wise multiplication across channels, the ones that remain are colored bright green and the nullified channels are brown and transparent. \label{fig_trl_high}}
\end{figure*}

Prior to being defined as the \textit{destructive interference} problem, research in MTL was only implicitly attempting to solve it. For example, statistically analyzing task relationships \cite{mtl_msd} and infusing the findings as prior knowledge helps combining tasks that benefit from each others learning process. Or design choices for a complete multilevel output architecture \cite{kokkinos2017ubernet} with structured sharing can rest on firm domain experience. This allows Ubernet \cite{kokkinos2017ubernet} to create shared features between low, mid and high level tasks at the correct levels within the model. Prior knowledge can be utilized in the final branch-output of a branched MTL architecture as well, where carefully designing the task-specific branches improves the model's predictive performance \cite{ranjan2019hyperface}. However, the above statistical analyses, architecture and branch design choices rely on prior knowledge, which is often unavailable or expensive to obtain. We mitigate the issue of obtaining prior knowledge, by introducing a task-routing mechanism allowing tasks to have separate in-model data flows. In this way, by enforcing a structured random solution we allow tasks to forge their own beneficial way of sharing.

We empirically verify our routing mechanism's positive influence on the task number scalability capacity through gradually increasing the number of tasks performed by a single model in our experiments. Starting from a relatively small number of tasks namely four in the Zappos-50K dataset, we scale up to 312 tasks in the UCSD-Birds dataset \cite{wah2011caltech}. To distinguish this setup from regular MTL, we introduce Many Task Learning (MaTL) as a special case of MTL where more than 20 tasks are performed. For MTL we show competitive performance with a small task count, and further proceed beyond the capabilities of existing methods to achieve state-of-the-art performance on the complete set of possible tasks in the UCSD-Birds dataset in an MaTL context. 

In this paper, we identify the following primary contributions:

\begin{itemize}
	\item We present a scalable MTL technique for exploiting cross-task expertise transferability without requiring prior domain knowledge.
	\item We enable structured deterministic sampling of multiple sub-architectures within a single MTL model.
	\item We forge task relationships in an intuitive non-parametric manner during training without requiring prior domain knowledge or statistical analysis.
	\item We apply our method to 5 classification and retrieval datasets demonstrating its effectiveness and performance gains over strong baselines and state-of-the-art approaches.
\end{itemize} 

\section{Related Work}

As our method draws inspiration from feature-wise transformation, architecture search and regularization works, this section is structured to cover those domains. As such, first we explain the ideas behind MTL and its possible variations. After that we link to related ideas in modulation and feature-wise transformations in an MTL context, and we complete the related work discussion by distinguishing our method from existing regularization and architecture search methods.

Multi-task learning (MTL) \cite{bakker2003task, Caruana1997, thrun1996learning} is a learning paradigm which seeks to improve the generalization performance of machine learning models optimizing for more than one task. Caruana \cite{Caruana1997} further describes MTL as a mechanism for improving generalization properties by leveraging the domain-specific information contained in the training signals of related tasks. As such, in MTL the goal is to jointly perform experiments over multiple tasks and improve the learning process for each of them. Whether these experiments are optimized for simultaneously or in an incremental fashion, categorizes MTL approaches in either \textit{symmetric} or \textit{asymmetric} \cite{xue2007multi}.

Asymmetric MTL relies on using knowledge from solving auxiliary tasks in order to improve the performance on one main target task. This formulation bears a resemblance to transfer learning \cite{survey_tl}. One key distinction between them is that in asymmetric MTL the auxiliary tasks are learned simultaneously with the main task, while in transfer learning they are learned independently \cite{zhang2017survey}. In our work we focus on symmetric MTL.

Symmetric MTL, unlike Asymmetric MTL, aims to improve the performance of all tasks simultaneously. It leverages the fact that some tasks are correlated (co-dependent) and by learning their estimators jointly under a unified representation, the transferability of expertise between tasks is exploited to the maximal benefit of all \cite{xue2007multi}. Zhang et al. introduced such a symmetric approach named Multi-task Relationship Learning (MTRL) \cite{zhang2014regularization} which regularizes the parallel learning of multiple tasks and models their relationships in a non-parametric manner as a task covariance matrix. Many other symmetric approaches \cite{misra2016cross, zhang2016deep, liu2015multi, li2014heterogeneous, zhang2014facial, zamir2018taskonomy} have been developed in recent years. Permutations of utilizing different regularization strategies \cite{li2014heterogeneous}, multi-level sharing \cite{zamir2018taskonomy}, cross-layer parameter combinations \cite{misra2016cross} or meshes of all options \cite{sluice2019ruder} have been extensively tested, however they are vulnerable to noisy and outlier tasks which when introduced dramatically deteriorate performance. This occurs due to the low grade feature robustness and the initial assumption that all tasks positively influence each other's learning process \cite{zhang2017survey}. In our work we address the feature robustness issue by randomizing the sharing structure from the start of the training process and enforcing tasks to use alternate routes for their data-flow through the model. 

Both symmetric and asymmetric MTL approaches often rely on prior knowledge to help with architecture design, sharing options and task grouping \cite{ranjan2019hyperface, teichmann2018multinet, jou2016deep, mtl_msd}. If such knowledge is present it is a helpful resource for designing an MTL model. However, often times such knowledge is unavailable and requires domain expert analysis (e.g. handcrafting an MTL model for the Omniglot \cite{Lake1332} dataset needs thorough knowledge of ancient alphabets). For this reason developing MTL models without prior domain knowledge is crucial to real-world applications. A recent step in this direction is made by Liu et al.\cite{lu2017fully} who propose an adaptive MTL model that structurally groups tasks together. Evolutionary algorithms have also been shown to capture task relatedness and create sharing structures \cite{architecture_search_mtl}. A less architectural solution is proposed by Yang et al. \cite{yang2017deep} who use a factorized space representation to initialize and learn inter-task sharing structures at each layer in an MTL model. Most of these methods have firm constraints in terms of how a model should be defined, structured, or initialized. We propose an approach applicable to any deep MTL model with no structural adjustments, as we encapsulate the layer-wise parameter space. By controlling the data-flow instead of the structure, we do not affect the underlying model behavior and this broadens the usability scope of our method.

Resource consumption is becoming predominantly important in MTL models as it usually increases with the number of performed tasks. As our approach is not structure dependent, it has a very small memory and computational footprint.  A recent scalable approach to MTL using modulation modules for image retrieval was proposed by Xiangyun et al. \cite{modulation2018} where they successfully scale to performing 20 and 40 tasks. The trade-off between speed and memory size in \cite{modulation2018} shows only 15\% overhead in feed-forward passes. In our work we build on this approach and demonstrate competitive performance on 300+ tasks in a single model with minimal costs to the computational budget, which with current methods is either inefficient or impossible. 

PackNet \cite{mallya2018packnet} presents an idea related to our work in the sense that Mallya et al. use the fixed weights of an existing network to learn a new task with the same model. This is an intuitive and simple method for re-usability of backbone networks for performing additional tasks, however as the authors indicate it has the downside of not allowing the tasks to share and benefit from each other's learning process. In cases such as this, Adaptive Instance Normalization \cite{huang2017arbitrary} is an approach that is able to adapt the channel-wise mean and variances between two inputs with no learnable parameters. This offers a similar feature-wise transformation and does not suffer from the same issue as \cite{mallya2018packnet}, but has not been tested in an MTL scenario where the inputs are task specific representations.  

Convolutional neural fabrics \cite{saxena2016convolutional} is related to our work in terms of architecture search. Saxena et al. define 3D trellis that connect response maps from different layers and create a smaller, thinner specialized architecture. On the other, TR works in the MTL realm and allows us to pool from an exponentially large pool of subnetworks. 

Similarly, Dropout \cite{srivastava2014dropout} relates to our work as a form of regularization and co-adaptation prevention technique. In dropout, the dropped-out units change each time the Bernoulli vector is sampled, which adds a stochastic component to this technique further inhibiting unit co-adaptation. The same regularization building block is present in related approaches \cite{singh2016swapout, wan2013regularization, ghiasi2018dropblock, yamada2018shakedrop} in an STL scenario. TR allows for a more deterministic form of inter-task regularization in a symmetric MTL paradigm. Furthermore, Dropout can be applied and prove beneficial in combination with our method as it can provide general regularization and additional co-adaptation prevention during training.

\section{Task Routing}

Most MTL methods involve task specific and shared units as part of the MTL training procedure. Our method enables the units within the model's convolutional layers to have a consistent \textit{shared} or \textit{task-specific} role in both training and testing regimes. Figure \ref{fig_one} provides the intuition behind how the individual units are utilized throughout the set of tasks performed by the model. We achieve this behavior by applying a channel-wise task-specific binary mask over the convolutional activations, restricting the input to the following layer to contain only activations assigned to the task. Figure \ref{fig_trl_high} illustrates the masking process over the activations. Because the flow of activations does not follow its conventional route, i.e. it has been rerouted to an alternate one, we have named our method Task Routing (TR) and its corresponding layer the Task Routing Layer (TRL). By applying the TRL to the network we are able to reuse units between tasks and scale up the number of tasks that can be performed with a single model. 

The masks that enable task routing are generated randomly at the moment the model is instantiated and kept constant through the training process. These masks are created using a sharing ratio hyper-parameter $\sigma$ defined beforehand. The sharing ratio defines how many units are task specific, and how many are shared between tasks. The inverse of this ratio determines how many of the units are nullified. As such, the sharing ratio enables us to explore the complete space of sharing possibilities with a simple adjustment of one hyper-parameter. A sharing ratio of 0 would indicate that no sharing occurs within the network and each trainable unit is specific to a single task only, resulting in distinct networks per task. On the opposite side of the spectrum, a sharing ratio of 1 would make every unit shared between each of the tasks, resulting in a classical fully shared MTL architecture.

\subsection{Task Routing Mask Creation}

Task routing is performed by means of a feature-wise transformation of the unit activations in a convolutional layer with a conditional binary mask. As our system does not have any prior knowledge for the problem at hand, the masks are created randomly at the moment our model is instantiated. The resulting random structure is persistent through the training and testing periods as the masks are not trainable.

Having immutable masks is particularly useful for MaTL, in which the space of possible sharing strategies is very large. By enforcing a fixed sharing strategy from the start of the training process, the model can focus on training robust task-specific and shared units, as opposed to training units over an ever changing combination of tasks.

\subsection{Task Routing Layer}

\begin{figure}[h!]
	\includegraphics[width=\linewidth]{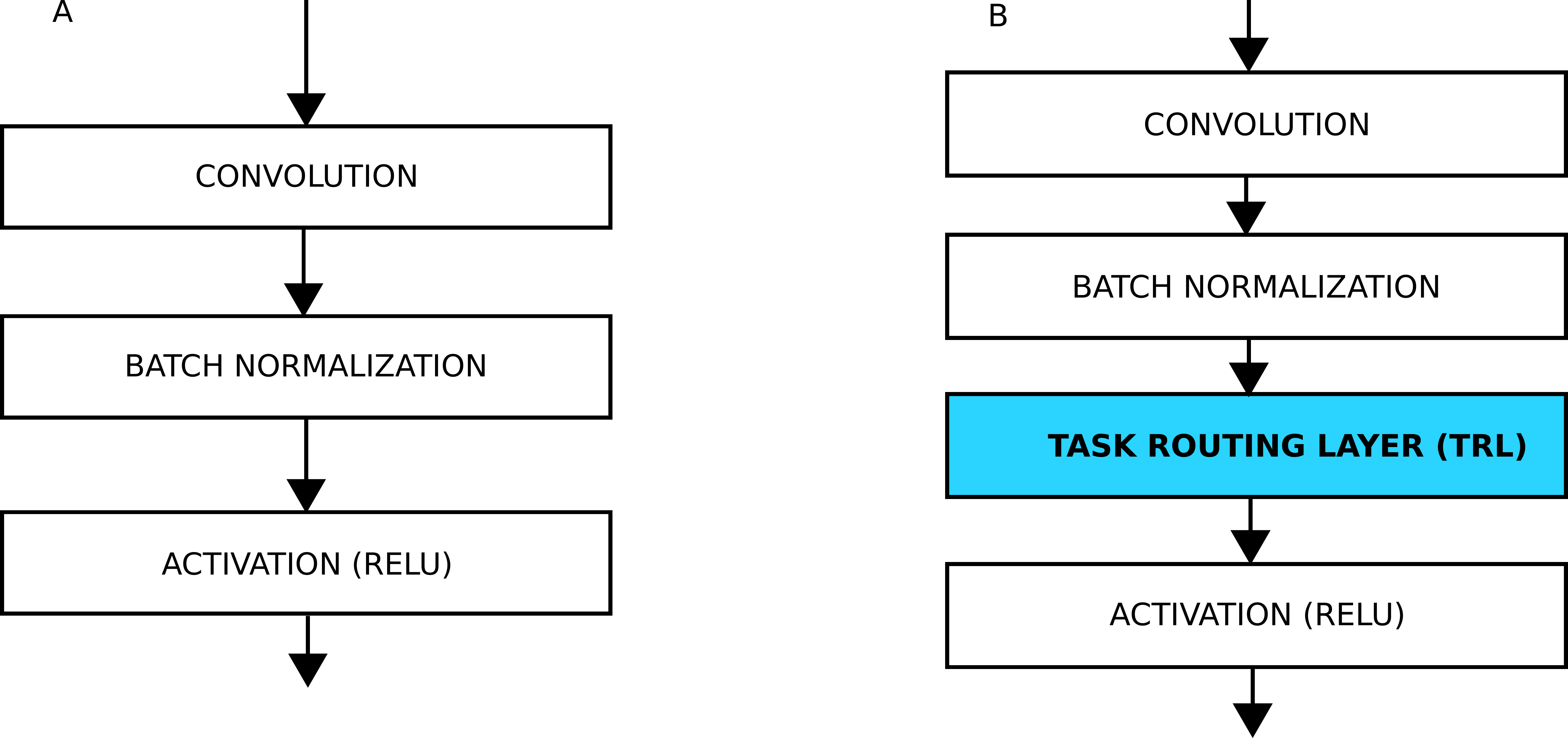}
	\caption{TRL placement (blue block) within a convolutional block. Section A (on the left) shows the convolutional block before adding the TRL, and Section B (on the right) after. \label{fig_placement}}
\end{figure}

We propose a new layer dubbed \textit{Task Routing Layer (TRL)} which contains a task specific binary mask $m_A \in \mathcal{Z}_2^C$ for the active task $A$ for over the input $X \in \mathcal{R}^{C \times H \times W}$ of a convolutional layer with dimensionality $[B\times C\times H\times W]$ where $C$ is the number of units,  $H$ the height and $W$ the width of the unit. For simplicity of notation we drop the H and W dimensions, as the mask is applied to an entire channel uniformly across the spatial dimensions. Applying this mask (see equation \ref{eq_tr_layer}) is akin to performing a conditional feature-wise transformation and generates a masked output which is then propagated to the next convolutional block. Figure \ref{fig_placement} shows the TRL placement within a convolutional block. Because the feature-wise transform applied to the convolutional output can affect the running mean and variance, the TRL is placed right after the batch normalization layer (if present).

\begin{equation}
TRL_{A}(X)= m_A \odot  X
\label{eq_tr_layer}
\end{equation}

\vspace{0.5cm}

During a forward pass a single specialized subnet is active, namely the subnet for the active task $A$. This is achieved by setting the active task for the TRLs. During our forward propagation we randomly sample one task from the pool of tasks. As the number of iterations required to traverse the dataset is usually much higher than the number of tasks, there exists a very small chance that a task is not optimized for within an epoch. This chance diminishes drastically with the training process spanning over multiple epochs. Even if we consider that a task has not been optimized for in an epoch, this is easily compensated for by the optimization of the other tasks that partly share the same group of units.

\begin{algorithm}[h]
	\caption{Training epoch for TRL}\label{alg_trl}
	\begin{algorithmic}[1]
		\Procedure{train}{X}
		\For{\texttt{ X in $X_{Train}$}} \Comment{Training loop}
		\State$A \gets sample(task\_set)$
		\State set\_active\_task($A$)
		\State forward($X$)
		\EndFor
		\EndProcedure
	\end{algorithmic}
\end{algorithm}

The training operational flow of our method is illustrated in algorithm \ref{alg_trl}. As we iterate through the training set, with each sampled mini-batch we change the currently active task. Setting the currently active task is a global change in the framework, so the TR workflow does not affect existing ways of propagation and training. This property makes TR simple to integrate in existing projects. As a global variable, the active task affects the applied mask in all TRLs in the model and navigates the routed activations to the task specific classifier.

In the forward pass of the input, the TRL applies the selected mask for the active task uniformly across the spatial dimensions of the entire input batch. As demonstrated in algorithm \ref{alg_forward} we perform the feature-wise transformation in the forward pass of the TRL over the convolutional layer's output. Figure \ref{fig_trl_high} illustrates how the task routing is performed and how the channels are nullified by the active mask.

\begin{algorithm}[h]
	\caption{Forward pass for TRL}\label{alg_forward}
	\begin{algorithmic}[1]
		\Procedure{forward}{X}
		\State $m_A \gets M[A]$ \Comment{$M \gets set of masks$}
		\State $out \gets m_A \odot X$ \Comment{Across all channels}
		\State \textbf{return} $out$ \Comment{The masked output}
		\EndProcedure
	\end{algorithmic}
\end{algorithm}

\subsection{Complexity}

Our model only adds a minimal number of additional parameters compared to the hard shared MTL approaches \cite{Caruana1997} or cross-stitch networks \cite{misra2016cross}, and has a significantly lower parameter count compared to similar architecture search approaches \cite{saxena2016convolutional}. The models we define in our experimental setup contain the task routing layers after each convolutional layer. This way, the number of additional parameters of the TRL is directly linked and proportional to the number of convolutional layers, units and channels.

Increasing the number of tasks, without appending a distinct embedding per task results in an negligible parameter count increase. However, heuristically we determined that having a separate embedding space per task increases standalone task performance. Because of this, the majority of the additional parameters in our models come from the wide task specific branches, rather than the TRLs. 

\section{Experimental Design}

Our experiments are designed to test and validate the contributions presented in this work. We evaluate our approach on multiple classification tasks, comparing to strong baselines and state of the art approaches. For this we consider a variety of datasets ranging from gray-scale proof of concept datasets (FashionMNIST), to attribute rich real world problems (UCSD-Birds) and multi-attribute based face dataset (CelebA). Through the CelebA and UT-Zappos50K dataset we also measure the effects of our method on the \textit{destructive interference} problem identified in \cite{modulation2018}. 

\subsection{Datasets}

\textbf{FashionMNIST} \cite{xiao2017/online} and \textbf{CIFAR-10} \cite{cifar} constitute the proof of concept part of our experimental design as they are well established benchmarks and provide simple indication of how different hyper-parameter setups tend to affect the method. For both dataset we define ten binary classification tasks and evaluate the accuracy, precision, and recall scores.

\textbf{UCSD Birds} \cite{wah2011caltech} is a dataset that provides 11.788 bird images over 200 bird species with 312 binary attribute annotations. For state of the art comparison, we compare on ten target attributes obtained with spectral clustering using the FSIC as the similarity measure \cite{mtl_msd}. As we incrementally increase the  selected number of attributes we define sets of 50, 100, 200, and 312 binary classification tasks for each of them. For this dataset the training and testing set have equal sizes and distributions. The attributes are sampled according to the ten attribute selection in \cite{mtl_msd} for the 10 task experiment and in order of the original annotation file for the rest of the experiments.

\textbf{CelebA} \cite{liu2018large} consists of more than 200,000 face images with binary annotations on 40 face attributes related to age, expression, decoration, etc. The first 10 attributes from \cite{modulation2018} are selected for the 10 task experiment as more related to face appearance. We additionally report the results on 40 attributes to compare our approach to \cite{modulation2018} in a classification setting.

\begin{figure*}
	\includegraphics[width=\textwidth]{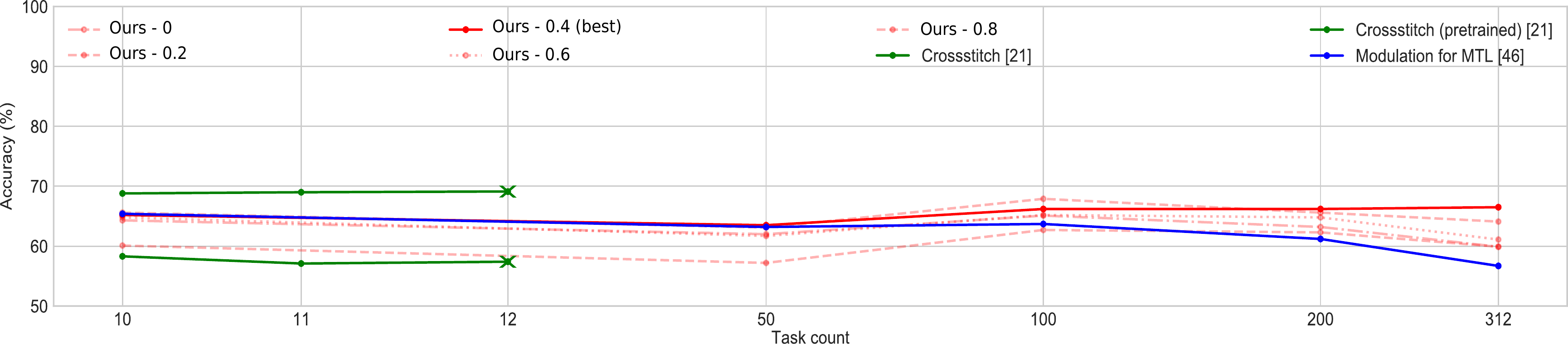}
	\caption{Accuracy comparison on the UCSD-Birds dataset on 10, 50, 100, 200, 312 task between our method (in red) with $\sigma=[0, 1]$, cross-stitch networks \cite{misra2016cross} (in green) and modulation for MTL \cite{modulation2018} (in blue). Cross-stitch networks scale up to 12 tasks, where as modulation for MTL and our approach fit to the full number of tasks. The best performing sharing ratio $\sigma=0.4$ is set in strong red, and the other $\sigma$ values in light red. \label{fig_acc_plot}}
\end{figure*}

\textbf{UT-Zappos50K} \cite{semjitter} is a large shoe dataset consisting of more than 50,000 catalog images collected from the web. This dataset contains four attributes of interest for our experiment, namely shoe type, suggested gender, height of the heel, and the shoe closing mechanism. As defined in \cite{modulation2018}, we define 4 classification tasks for a small scale test of our approach over a real world dataset using the identical train, validation, and test splits from \cite{veit2017conditional, modulation2018}.

\subsection{Multi-task Setup}

FashionMNIST and CIFAR10 power a toy problem experiment where we develop an intuition of how our method functions. We choose these datasets as they are well established and balanced benchmarks, for which little domain knowledge is necessary to interpret the results and draw conclusions. For each dataset we perform 10 binary classification tasks using the model presented by Xiao et al. \cite{xiao2017/online} for FashionMNIST and a VGG-16 network for CIFAR-10. In a similar manner, the Zappos50K dataset provides a highly related balanced dataset with four well described tasks on which we evaluate our method's performance in a small scale MTL context. 

To evaluate our method in an MaTL context, we perform experiments on the CelebA and UCSD-Birds datasets. For the CelebA experiment we ran experiments with an increasing number of tasks, starting from 10 and ending with 40 tasks. The purpose of these experiments is to observe the difference in performance with \cite{modulation2018} and explore how adding additional tasks affects the learning process and performance. This CelebA experiment uses the VGG-16 model trained from scratch with batch normalization as a feature extraction platform and branches out to as many classification branches as there are tasks. Each classification branch is task specific and has an independent embedding space.

With 312 possible attributes related to bird appearance, the UCSD-Birds dataset presents a unique opportunity to explore the task-wise scalability properties of our method. Due to memory-constraints this experiment uses the VGG-11 model with batch normalization trained from scratch as a feature extraction platform. To ensure a fair comparison with cross-stitch networks \cite{misra2016cross} we evaluate this method in a pre-trained and trained from scratch setting where applicable.

Over all datasets we perform experiments with the full range of values for the sharing ratio $\sigma$. A value of 0 means that every task gets a distinct subnetwork and no sharing occurs between tasks. A value of 1 means that the full model structure is shared between all task which results in a hard shared MTL setup.

\begin{table*}[h]
	\centering
	\caption{Average scores on the on FashionMNIST, CIFAR-10, Zappos50K and CelebA (10 and 40 Tasks) on the complete dataset with the complete sharing ratio scope $\sigma=[0, 1]$. Because for $\sigma=0$ no sharing occurs and for $\sigma=1$ our approach reverts to hard shared MTL, we group them separately. The overall best performing approach across all datasets is highlighted in gray and the best approaches per dataset are in bold. Fields marked with \textit{n/a} signify experiments for which the method could not scale to the task count.}
	\label{tab_full}
	\begin{adjustbox}{max width=\linewidth}
		\begin{tabular}{l|lll|lll|lll|lll|lll}
			\hline
			Dataset	& \multicolumn{3}{c|}{FashionMNIST} & \multicolumn{3}{c|}{CIFAR-10} & \multicolumn{3}{c|}{Zappos50K} & \multicolumn{3}{c|}{CelebA} & \multicolumn{3}{c}{CelebA (Full)} \Tstrut\\ 
			\hline
			Number of Tasks& \multicolumn{3}{c|}{10} & \multicolumn{3}{c|}{10} & \multicolumn{3}{c|}{4} & \multicolumn{3}{c|}{10} & \multicolumn{3}{c}{40} \Tstrut \\ 
			\hline
			Approach & Accuracy & Precision & Recall & Accuracy & Precision & Recall & Accuracy & Precision & Recall& Accuracy & Precision  & Recall & Accuracy & Precision  & Recall \Tstrut \\
			\hline
			Cross-stitch \cite{misra2016cross}&\textbf{98.1} & 91.4 & 86.1 & 98.5 & 91.6 & 85.9& 84.7 & 82.2 & 81.8& 71.5 & 68.0 & 67.0 &n/a&n/a&n/a \Tstrut \\
			Modulation \cite{modulation2018}& 96.9 & 91.0 & 80.1&63.2&57.4&53.2&63.7&60.4&59.8& 71.9 & 70.2 & 69.4 & 64.1 & 61.0 & 60.4 \\
			\hline
			Ours $\sigma=0$     & 97.8 & 91.9 & 85.5 & 96.5 & 89.1 & \textbf{87.8} & 88.3 & 83.1 & 83.2& 70.1 & 68.0  & 67.4 & 63.1 & 60.8 & 60.0 \Tstrut \\
			Ours $\sigma=1$ & 97.4 & 91.1 & 85.1 & 98.1 & 88.0 & 85.6 & 79.2 & 77.1 & 75.3 & 69.9 & 67.2  & 66.8  & 62.2 & 58.0 & 56.4 \\
			\hline
			\rowcolor{Gray}
			Ours $\sigma=0.2$   & 97.8 & \textbf{92.2} & \textbf{85.7} & \textbf{99.0} & \textbf{92.1} &85.2    & \textbf{89.5} & \textbf{85.2} & \textbf{83.4} & \textbf{73.2} & \textbf{71.4} & \textbf{70.8} & 63.1 & 60.8 & 60.0 \Tstrut \\
			Ours $\sigma=0.4$   & 97.6 & 92.0 & 84.2 & 96.9 & 92.0 & 87.5 & 88.1 & 84.3 & 82.8 & 73.0 & 71.4  & 70.2 & 62.0 & 59.4 & 58.2 \\
			Ours $\sigma=0.6$   & 97.1 & 91.1 & 80.4 & 96.0 & 90.3 & 84.3 & 87.4 & 82.2 & 82.0 & 72.7 & 71.0  & 69.6 & \textbf{65.3} & \textbf{62.4} & \textbf{61.8} \\
			Ours $\sigma=0.8$   & 96.8 & 91.0 & 78.0 & 94.8 & 88.2 & 81.1 & 83.2 & 81.4 & 78.9 & 71.4 & 70.1  & 69.1 & 64.0 & 61.1 & 60.2 \\
			\hline
		\end{tabular}
	\end{adjustbox}
\end{table*}

\subsection{Implementation Details}

In all classification settings we perform binary classification tasks over the attributes. Each attribute is considered a binary classification task and has its own equal-sized embedding space. We append the TRL after each convolutional layer in our models and randomly initialize the routing maps as the model is instantiated. For all our experiments we use existing model architectures (VGG-11, VGG-16 \cite{simonyan2014very} and Resnet50 \cite{he2016deep}) with their default architecture settings. For all datasets we use a batch size of 64 images with normalization by the dataset mean. For the UCSD-Birds dataset we explore both trained from scratch and pretrained VGG-11 models with horizontal flipping because of the small size of the training set. We use Stochastic Gradient Descent with a learning rate of 0.01 and momentum of 0.5 with which our method usually converges after 35 epochs. \footnote{The source code will be released on Github upon acceptance.}

\subsection{Evaluation criteria}

For evaluating the performance of our method we track the accuracy, precision and recall. We added precision to our evaluation criteria because it is important to highlight how precise the model is, i.e. how many of the positive predictions are true positives. This gives insight into how precise and how robust our specialized subnets are the task specific representation is. Tracking recall gives a realistic measure how well the model has adapted to each of the tasks as it shows how much of the actual positive samples are covered. 

\begin{figure}[h]
	\includegraphics[width=\linewidth]{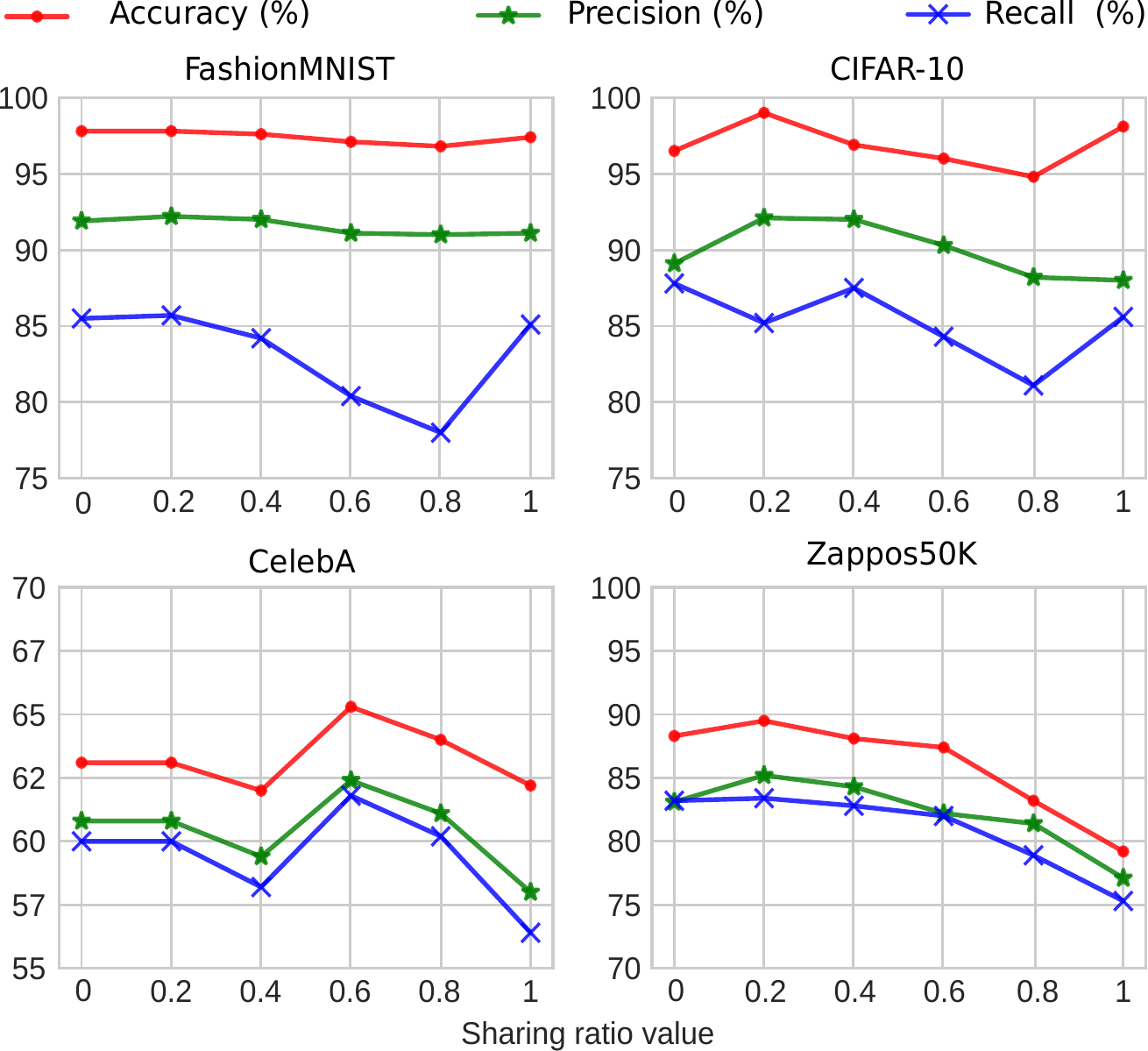}
	\caption{Effects of the sharing ratio $\sigma$ value on accuracy, precision and recall in Fashion-MNIST (top left), CIFAR10 (top right),  CelebA (bottom left) and Zappos50K (bottom right). Partially sharing units between tasks is beneficial to performance with sharing ratios $\sigma=[0.2, 0.6]$ compared to a fully shared network $\sigma=1$ or many distinct subnetworks without sharing $\sigma=0$. \label{fig_sigma_param}}
\end{figure}

\section{Results}

We evaluate our method on five datasets with experiments ranging from proof of concept (FashionMNIST, CIFAR10 and Zappos50K) to addressing the task count scalability properties (UCSD-Birds and CelebA). We report performance for the full scope of possible values for our method specific hyper-parameter $\sigma$, as compared to modulation for MTL \cite{modulation2018} and cross-stitch networks \cite{misra2016cross}. With $\sigma = [0,1]$ and a varying number of tasks per model we explore the complete space of sharing capabilities our method has to offer ranging from a distinct specialized subnet per task $\sigma=0$, to a completely shared structure when $\sigma=1$.

The experimental results for this comparison are reported in Table \ref{tab_full}. The results show that using our method we are able to efficiently use the units in a single model to fit and optimize for many tasks. Moreover, we surpass the performance of the modulation for MTL method \cite{modulation2018}, as well as the cross-stitch networks approach \cite{misra2016cross} over five datasets in an MTL/MaTL setup. This performance is shown in Figure \ref{fig_acc_plot}.

\begin{table*}[th!]
	\centering
	\caption{Average scores using the routing module over an increasing number tasks for the UCSD-Birds dataset and a sharing ratio of $\sigma=[0, 1]$. Because for $\sigma=0$ no sharing occurs and for $\sigma=1$ our approach reverts to hard shared MTL, we group them separately. Fields marked with \textit{n/a} signify experiments for which the method could not scale to the task count. The pretrained cross-stitch networks experiment is marked with a star (*). The overall best performing method is highlighted in gray and the best performing model per task setting is set in bold.}
	\label{tab_birds}
	\begin{adjustbox}{max width=\linewidth}
		\begin{tabular}{l|lll|lll|lll|lll|lll}
			\hline
			Dataset& \multicolumn{15}{c}{UCSD-Birds} \Tstrut\\
			\hline
			Number of tasks& \multicolumn{3}{c|}{10} & \multicolumn{3}{c|}{50}  & \multicolumn{3}{c|}{100} & \multicolumn{3}{c|}{200} & \multicolumn{3}{c}{312} \Tstrut\\ 
			\hline
			Approach & Accuracy & Precision & Recall & Accuracy & Precision & Recall & Accuracy & Precision & Recall & Accuracy & Precision & Recall & Accuracy & Precision & Recall \Tstrut \\
			\hline
			Cross-stitch \cite{misra2016cross}&58.3&55.6&54.2&n/a&n/a&n/a&n/a&n/a&n/a&n/a&n/a &n/a&n/a&n/a&n/a  \Tstrut\\
			Cross-stitch * \cite{misra2016cross}&\textbf{68.8}&\textbf{67.4}&\textbf{67.0} &n/a &n/a&n/a&n/a&n/a&n/a&n/a&n/a&n/a&n/a&n/a&n/a  \\
			Modulation \cite{modulation2018}&65.4&59.8&55.2&63.2&57.4&53.2&63.7&60.4&59.8&61.2&58.6 &57.3 &56.7& 51.8&50.2\\
			
			\hline
			Ours $\sigma=0$   &64.3&62.4&55.3&62.0&60.6&54.6&65.1&62.7&61.1&63.2&60.2&58.8&59.9&57.2&56.1 \Tstrut\\
			Ours $\sigma=1$ &62.3&57.4&51.8&58.6&56.8&54.2&60.7&58.1&57.8&60.0&58.6&56.8 &59.6&53.9&52.2 \\
			\hline
			Ours $\sigma=0.2$ &65.6&62.9&57.0&63.1&62.9&57.2&\textbf{67.8}&63.3&60.9&65.6&63.6 &63.2 &64.1& 61.6&60.2 \Tstrut \\
			\rowcolor{Gray}
			Ours $\sigma=0.4$ &65.1&62.7&55.9&\textbf{63.5}&\textbf{63.0}&\textbf{59.9}&66.2&\textbf{63.8}&\textbf{61.2}&\textbf{66.2}& \textbf{64.2}&\textbf{63.7}&\textbf{66.5}&\textbf{62.3}&\textbf{61.8}\\
			Ours $\sigma=0.6$ &64.9&62.1&54.8&61.7&59.9&59.0&65.2&60.9&59.5&64.8&62.0&59.8 &61.1&59.2&59.0\\
			Ours $\sigma=0.8$ &60.1&55.0&50.2&57.2&52.2&50.0&62.7&60.4&59.8&62.3&59.2&58.0 &59.9&55.1&54.2\\
			\hline
		\end{tabular}
	\end{adjustbox}
\end{table*}

An important feature of our method is its scalability regarding the number of tasks a single model can accommodate using the TRL. Table \ref{tab_birds} shows the relationship between the number of tasks, the sharing ratio coefficient $\sigma$ and accuracy, precision and recall as evaluation metrics. For sharing ratios $\sigma=0.2$ and $\sigma=0.4$ we can observe a consistent improvement on all three scores as the task count is increased in Table \ref{tab_birds}. Performance significantly drops with a sharing ratio of $\sigma=0.8$ as the model is closing in on a hard shared MTL architecture. In this case only 20\% of the units remain task specific after the routing. A sharing ratio of $\sigma=1$ means that every unit is used by every task and the reported performance is equal to that of a classical MTL hard shared approach (see Table \ref{tab_full} and \ref{tab_birds}). Figure \ref{fig_sigma_param} shows the effects of the sharing ratio parameter $\sigma$ over the evaluation metrics. 

For the CelebA we performed an experiment with ten tasks (smile, mouth-open, no-beard, mustache, sideburns, young, double-chin, chubby, are-eyebrows, high-cheekbone) and 40 tasks (the complete attribute set). We consistently witness a performance drop across all applicable approaches once the remaining 30 tasks are added to the task pool. From these results, a plausible conclusion is that it is more difficult to perform well on the additional 30 tasks. We suspect that this is due to a smaller sample number of positive instances in the training set for the additional tasks when used in a classification setting.

Considering the task count scalability of our method compared to competing approaches, Figure \ref{fig_acc_plot} illustrates the task count to performance relation for cross-stitch networks \cite{saxena2016convolutional}, modulation for MTL \cite{modulation2018}, a hard shared MTL baseline (our approach with $\sigma=1$) and our approach over the complete sharing space ($\sigma=[0,0.8]$). As cross-stitch networks require independent models per task between which unit sharing occurs, every additional task requires a complete model to be loaded into memory. Despite having good performance even with using a small VGG-11 network, it becomes impossible to fit this approach into memory with more than 12 tasks. If a more complex model is used, such as a ResNet \cite{he2016deep} or an Xception network \cite{DBLP:journals/corr/Chollet16a} the task count capacity of this approach diminishes significantly. On the other hand, our approach and the modulation for MTL are more parameter savvy and can fit more tasks. However, when comparing the performance of our approach to modulation for MTL we can observe a slight gain in performance for our approach as task count increases, whereas modulation for MTL drops in performance (see Figure \ref{fig_acc_plot}).

\section{Conclusion}

In this work, we have presented a method to efficiently perform a large number of classification tasks with a single model. The proposed method allows us to modify the default behavior of an MTL model and apply a conditional feature-wise transformation to the outputs of its convolutional layers. A strong point of our approach is that it does not require prior knowledge of the domain or the intertask relationships to achieve good performance in both regular MTL and MaTL settings.

At the core of our method is a layer dubbed the \textit{Task Routing Layer}, that can be inserted after any convolutional layer within a model's architecture with minimal effort and computational overhead. This layer contains task specific masks that allow for a single model to fit to many tasks within its parameter space. By passing the input through the mask we are training specialized subnetworks per task with much lower dimensionality compared to the main model. The dimensionality of the specialized subnets is dictated by the sharing ratio hyper-parameter, and they can be extracted and used in place of the complete model for a specific task.

Furthermore, the sharing ratio hyper-parameter $\sigma$ gives our method an additional degree of freedom when compared to other state-of-the-art approaches and baseline methods. The sharing ratio $\sigma$ allows us to be flexible with the task routing design without changing the underlying architecture which proves beneficial in MTL and MaTL. As an universal solution to most of the problems does not exist, we offer a simple way to explore all sharing possibilities the model has to offer. Finally, our approach offers an intuitive and easy to implement mechanism to get more out of existing models.

\section{Acknowledgements}
This research is supported by the VISTORY project.

{\small
	\bibliographystyle{ieee}
	\bibliography{egbib}
}

\end{document}